\documentclass[letterpaper,10pt]{article}
\usepackage{arxiv}
\renewcommand{\shorttitle}{MyoAcross}
\renewcommand{\headeright}{}
\renewcommand{\undertitle}{}
\date{}
\usepackage{amsmath,amssymb}
\usepackage{graphicx}
\usepackage{booktabs}
\usepackage{array}
\usepackage{cite}
\usepackage[hidelinks]{hyperref}
\graphicspath{{figures/}}
\hypersetup{pdfauthor={Bozhao Wang, Yinglei Zhu, Zhengqing Li, Zhiyuan Zhang, Weiming Kang, Yiheng He, Xinjun Liu, Senhao Hou, Huichan Zhao, Xiaoqiang Tang},pdftitle={MyoAcross: A Framework for Scalable Individualized Musculoskeletal Simulation with a Generalist Morphology-Conditioned Policy}}

\makeatletter
\renewcommand{\@maketitle}{%
  \vbox{%
    \hsize\textwidth
    \linewidth\hsize
    \@toptitlebar
    \centering
    {\LARGE\scshape \@title\par}
    \@bottomtitlebar
    \vspace{10pt}
    {\normalfont\normalsize \@author\par}
    \vspace{12pt}
  }%
}
\makeatother

\usepackage{chapterbib}
\usepackage{float}
\begin{document}
\raggedbottom
\title{MyoAcross: A Framework for Scalable Individualized Musculoskeletal Simulation with a Generalist Morphology-Conditioned Policy}
\newcommand{\MyoAuthorBlock}{%
\normalfont Bozhao Wang\textsuperscript{1,\dag}, Yinglei Zhu\textsuperscript{1,\dag}, Zhengqing Li\textsuperscript{1},\\
Zhiyuan Zhang\textsuperscript{1}, Weiming Kang\textsuperscript{1}, Yiheng He\textsuperscript{1}, Xinjun Liu\textsuperscript{1,2},\\
Senhao Hou\textsuperscript{1,*}, Huichan Zhao\textsuperscript{1,2,*}, Xiaoqiang Tang\textsuperscript{1,2,*}\\[12.5pt]
{\small
\textsuperscript{1}State Key Laboratory of Tribology in Advanced Equipment, Department of Mechanical Engineering,\\
Tsinghua University, Beijing 100084, China.\\[5.5pt]
\textsuperscript{2}Beijing Key Laboratory of Transformative High-end Manufacturing Equipment and Technology,\\
Tsinghua University, Beijing 100084, China.\\[9.5pt]
\textsuperscript{\dag}These authors contributed equally. \textsuperscript{*}Corresponding authors.\\[5.5pt]
Senhao Hou: \href{mailto:housenhao@tsinghua.edu.cn}{housenhao@tsinghua.edu.cn};\\
Huichan Zhao: \href{mailto:zhaohuichan@mail.tsinghua.edu.cn}{zhaohuichan@mail.tsinghua.edu.cn};\\
Xiaoqiang Tang: \href{mailto:tang-xq@tsinghua.edu.cn}{tang-xq@tsinghua.edu.cn}.
}}
\author{\MyoAuthorBlock}
\maketitle
\begingroup
\renewcommand{\thefootnote}{}
\footnotetext{This work was supported by the National Natural Science Foundation of China
(Grant No.\ 52575024) and the Fundamental and Interdisciplinary Disciplines
Breakthrough Plan of the Ministry of Education of China (Grant No.\ JYB2025XDXM208).}
\endgroup
\thispagestyle{plain}
\begin{abstract}
Musculoskeletal simulations powered by deep reinforcement learning (DRL) support human-robot interaction design and rehabilitation engineering, and they typically rely on a generic template model. However, differences in skeletal proportions and body mass alter dynamics, limiting the fidelity of generic models and the reuse of policies trained on them. Scaling individualized simulations requires a framework that connects model construction from accessible inputs with a generalist policy that transfers without fine-tuning. We present MyoAcross, which integrates skeletal-guided individual model construction, generalist morphology-conditioned policy learning, and virtual model augmentation for full-body muscle-driven control. Across 31 held-out individual models, MyoAcross achieves 99.24\% completion with a 53.70\,mm tracking error, compared with 8.32\% completion for a baseline policy trained only on the generic model. Ablation studies show that morphology conditioning improves tracking accuracy while virtual model augmentation broadens generalization. Across input modalities, the pretrained policy demonstrates generalizability to models constructed from body shape (paired with mass), monocular images (paired with height and mass), and marker-based motion capture (paired with mass), making MyoAcross scalable. For biomechanical assessment, MyoAcross improves agreement with measured ground reaction forces and surface electromyography over the generic model on the external Gait120 dataset. These results establish MyoAcross as a practical framework for scalable individualized musculoskeletal simulation. It is suited for systematic cross-individual applications, such as exoskeleton DRL assistance policy design.

\end{abstract}
\section{Introduction}
\label{sec:introduction}

\begin{figure*}[!t]
  \centering
  \includegraphics[width=\linewidth]{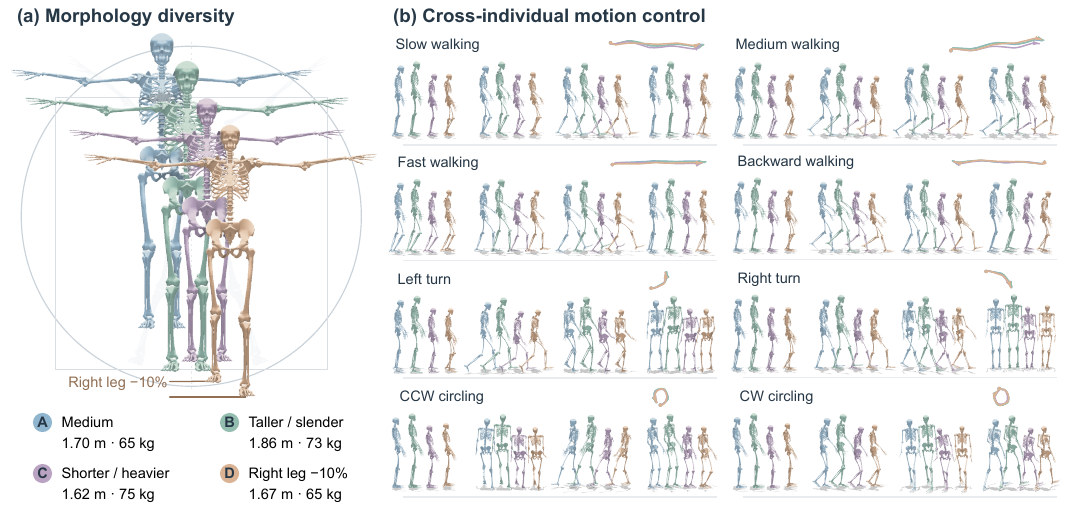}
  \caption{\textbf{Full-body muscle control across individual musculoskeletal models.}
  (a) Models with different body dimensions and masses; model D has 10\% shortening of the right thigh and shank.
  (b) Motion sequences generated by a pretrained generalist policy for the 4 models across 8 locomotion categories.
  Colors identify the models, and curves indicate movement paths. CW, clockwise; CCW, counterclockwise.}
  \label{fig:teaser}
\end{figure*}

Musculoskeletal simulations provide a powerful computational approach for
investigating human movement, analyzing neuromuscular function, and developing
human-robot interaction and rehabilitation
technologies~\cite{seth2011opensim,holzbaur2005model,luo2024exoskeleton}.
By simulating musculotendon and multibody dynamics, they enable the study of
internal biomechanical quantities that are difficult to measure directly, such
as muscle activations, forces, and joint moments, as well as responses to
external loads.
Because differences in skeletal proportions and body mass alter full-body
dynamics, biomechanical fidelity depends critically on how accurately the
underlying model represents each individual~\cite{zajac1989muscle}.
Furthermore, scaling individualized musculoskeletal simulation requires not only
accessible model construction and a GPU-accelerated simulation backend, but also
a generalist policy operating without fine-tuning.

\begin{table*}[!t]
  \centering
  \caption{Comparison of Representative Musculoskeletal Simulation Frameworks}
  \label{tab:framework-comparison}
  \small
  \setlength{\tabcolsep}{4pt}
  \renewcommand{\arraystretch}{1.18}
  \begin{tabular}{@{}>{\raggedright\arraybackslash}m{31mm}>{\raggedright\arraybackslash}m{27mm}>{\raggedright\arraybackslash}m{25mm}>{\centering\arraybackslash}m{23mm}>{\centering\arraybackslash}m{18mm}>{\centering\arraybackslash}m{27mm}@{}}
    \toprule
    \textbf{Framework} & \textbf{Physics\newline Backend} & \textbf{Input\newline Modalities} & \textbf{Individual\newline Model\newline Construction} & \textbf{Closed-Loop\newline DRL} & \textbf{Cross-Individual\newline Generalist Policy} \\
    \midrule
    OpenSim~\cite{seth2011opensim} & Simbody\newline (CPU) & Markers & $\checkmark$ & $\times$ & $\times$ \\
    AddBiomechanics~\cite{werling2023addbiomechanics} & NimblePhysics\newline (CPU) & Markers & $\checkmark$ & $\times$ & $\times$ \\
    SCONE~\cite{geijtenbeek2019scone,sconeRLDocumentation} & OpenSim / Hyfydy\newline (CPU) & Pre-built models & Imported models & $\checkmark$ & $\times$ \\
    MyoSuite~\cite{caggiano2022myosuite} & MuJoCo\newline (CPU / GPU$^{\dagger}$) & Generic templates & $\times$ & $\checkmark$ & $\times$ \\
    MuscleMimic~\cite{li2026musclemimic} & MuJoCo Warp\newline (GPU) & Generic template & $\times$ & $\checkmark$ & $\times$ \\
    MS-Emulator~\cite{wei2026msemulator} & MuJoCo Warp\newline (GPU) & Generic template & $\times$ & $\checkmark$ & $\times$ \\
    \midrule
    \textbf{MyoAcross (Ours)} & \textbf{MuJoCo Warp\newline (GPU)} & {\raggedright\textbf{SMPL body shape, image, markers}\par} & $\checkmark$ & $\checkmark$ & $\checkmark$ \\
    \bottomrule
  \end{tabular}
  \vspace{2pt}
  \begin{flushleft}
    Entries refer to native capabilities; cross-individual policy reuse requires no fine-tuning.\\
    $^{\dagger}$GPU support is available through backend extensions.
  \end{flushleft}
\end{table*}

Existing musculoskeletal simulation frameworks address distinct subsets of
these requirements (Table~\ref{tab:framework-comparison}).
Classical biomechanics platforms such as
OpenSim~\cite{seth2011opensim,opensimScaleTool},
AddBiomechanics~\cite{werling2023addbiomechanics}, and
SCONE~\cite{geijtenbeek2019scone}---together with imaging-based model builders
such as nmsBuilder~\cite{valente2017nmsbuilder} and reinforcement learning
tuning tools~\cite{berman2024personalizing}---support individual model
construction, parameter calibration, and movement analysis, but rely on
CPU-based simulation, laboratory marker protocols, medical imaging, or
electromyography (EMG) trials, and lack a GPU-accelerated deep reinforcement
learning (DRL) interface and a closed-loop generalist full-body policy.
Meanwhile, parametric body-recovery methods such as the anatomically informed
model SKEL and Human Skeleton and Mesh Recovery (HSMR) extract skeletal
geometry from Skinned Multi-Person Linear (SMPL)~\cite{loper2015smpl} body shape or monocular
images~\cite{keller2023skel,xia2025hsmr,mahmood2019amass}; although these
modalities provide a rich data ecosystem and accessible inputs, such methods do
not produce dynamically calibrated full-body musculoskeletal models with a
unified control interface.
On the control side, DRL frameworks have progressed from CPU-based partial-body
benchmarks such as MyoSuite~\cite{caggiano2022myosuite} and single-model
imitation~\cite{lee2019scalable,feng2023musclevae} to GPU-accelerated full-body
platforms such as MuscleMimic~\cite{li2026musclemimic} and
MS-Emulator~\cite{wei2026msemulator}, while rehabilitation and cross-morphology
studies have explored prosthetic gait~\cite{devree2021deep,ogum2024walking},
simplified planar arms~\cite{crowder2022learning}, and policies over parametric
or predefined morphologies~\cite{park2022gaitnet,kim2025freemusco,an2025arnold}.
However, as summarized in Table~\ref{tab:framework-comparison}, GPU-accelerated
full-body DRL frameworks remain centered on a generic template model,
whereas cross-individual muscle policies have not been unified with accessible
model construction and systematic generalization and biomechanical evaluation.

To bridge this gap, we present MyoAcross, a GPU-accelerated framework for
scalable individualized musculoskeletal simulation that integrates
skeletal-guided individual model construction, generalist morphology-conditioned
policy learning, and virtual model augmentation (Fig.~\ref{fig:teaser}).
Skeletal-guided regional scaling and mass calibration map target skeletal
measurements and body mass onto a unified 416-muscle full-body control
topology, accommodating inputs from SMPL body shape (paired with mass),
accessible monocular images (paired with height and mass), or marker-based
motion capture (paired with mass) for scalable deployment.
Model-derived morphology features condition the generalist policy on the
dynamics of the active individual model, while virtual model augmentation
broadens physical-parameter variation during training through structured,
correlated sampling of regional scale factors and body mass.
Together, these components enable the generalist policy to track full-body
reference motions across diverse individual models without fine-tuning.
Further experiments isolate the contributions of morphology conditioning and
virtual model augmentation through ablation studies and demonstrate improved
agreement with external human ground reaction force (GRF) and surface
electromyography (sEMG) measurements.

\begin{figure*}[!t]
  \centering
  \includegraphics[width=\linewidth]{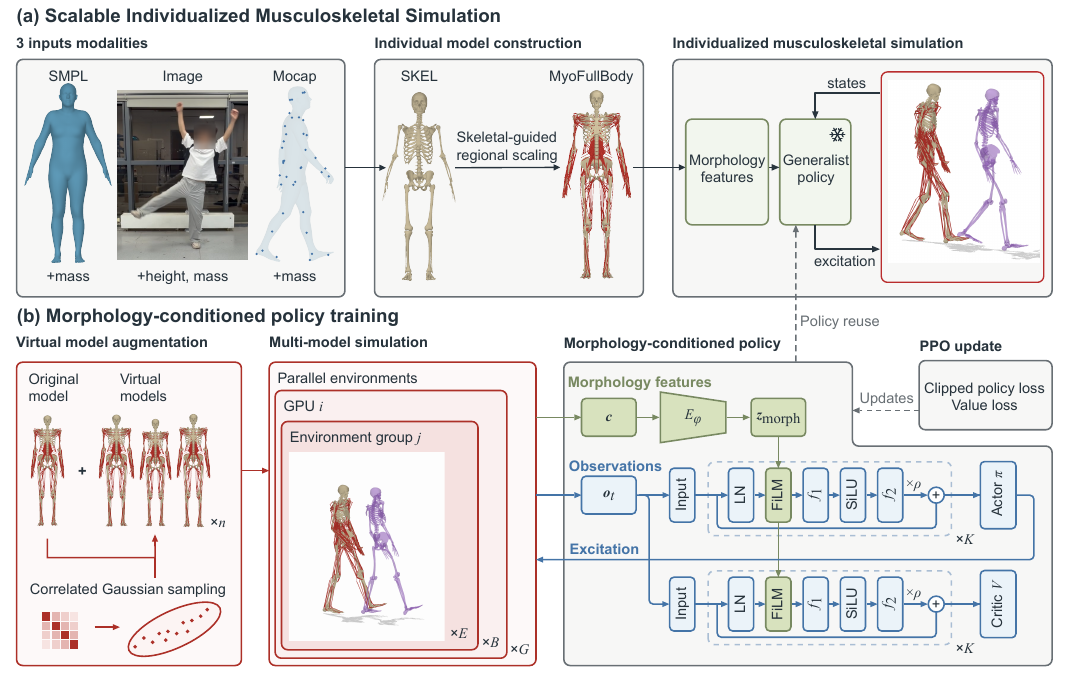}
  \caption{\textbf{Overview of the MyoAcross framework.}
  (a) Individualized musculoskeletal simulation from SMPL body shape, a monocular image (with height calibration), or marker-based motion capture. SKEL-derived skeletal measurements guide regional scaling of the MyoFullBody template, and morphology features $\boldsymbol{c}_i$ condition a frozen pretrained generalist policy (snowflake) to track reference motions $\mathcal{R}_i$ (purple skeletons).
  (b) Generalist policy training with virtual model augmentation in parallel environments, where feature-wise linear modulation (FiLM) conditions actor and critic networks under proximal policy optimization (PPO) updates; the dashed arrow denotes policy reuse in (a). The pictured author consents to publication.}
  \label{fig:framework}
\end{figure*}

Our main contributions are:
\begin{enumerate}
\item \textbf{A framework for scalable individualized musculoskeletal simulation.}
We connect skeletal-guided model construction from SMPL body shape, accessible
monocular images, and marker-based motion capture to a GPU-accelerated
416-muscle simulation backend, enabling a generalist policy to control diverse
individual models without fine-tuning.
\item \textbf{Generalist morphology-conditioned policy with virtual model augmentation.}
We show that policies trained on a generic model fail to control individual
models, and establish through ablation studies that model-derived morphology
conditioning and virtual model augmentation provide complementary gains in
tracking accuracy and cross-individual generalization.
\item \textbf{Comprehensive generalization, sensitivity, reproducibility, and external biomechanical validation.}
We evaluate generalization to unilateral lower-limb shortening and accessible
image-derived models, quantify sensitivity to input perturbations, verify
reproducibility across independent training runs, and demonstrate improved GRF
and sEMG agreement over generic and globally scaled models on the external
Gait120 dataset.
\end{enumerate}


\section{Methods}
\label{sec:methods}

MyoAcross integrates skeletal-guided individual model construction,
generalist morphology-conditioned policy learning, and virtual model
augmentation for scalable individualized musculoskeletal simulation
(Fig.~\ref{fig:framework}). Below, we formulate the control problem, detail the
framework components, and describe the experimental protocol.

\subsection{Control formulation}
We formulate muscle-driven reference-motion tracking as a high-dimensional
control problem for an individual musculoskeletal model whose coupled muscle
activation and multibody dynamics are governed by model-specific physical
parameters~\cite{seth2011opensim}. For model $i$, parameters
$\boldsymbol{\eta}_i$ specify geometry, mass distribution, inertia, and
musculotendon properties; the state comprises generalized coordinates
$\boldsymbol{q}$, velocities $\boldsymbol{v}$, and muscle activations
$\boldsymbol{a}\in[0,1]^{416}$, driven by muscle excitations
$\boldsymbol{e}\in[0,1]^{416}$.

For musculotendon unit $j$, excitation governs activation via first-order
dynamics, and resulting force depends on activation and model-specific
properties~\cite{millard2013flexing,mujocoMuscleModel}:
\begin{equation}
\begin{aligned}
\dot a^{(j)}&=\frac{e^{(j)}-a^{(j)}}{T^{(j)}(e^{(j)},a^{(j)})},\\
F^{(j)}(\boldsymbol{q},\boldsymbol{v},\boldsymbol{a};\boldsymbol{\eta}_i)
&=F_{\max}^{(j)}(\boldsymbol{\eta}_i)\bigl[
a^{(j)}f_L(\ell_i^{(j)})f_V(\nu_i^{(j)})\\
&\qquad\qquad\qquad+f_P(\ell_i^{(j)})\bigr].
\end{aligned}
\label{eq:muscle-activation-force}
\end{equation}
Here, $T^{(j)}$ is the effective activation/deactivation time constant,
$F_{\max}^{(j)}$ is maximum isometric force, and $f_L$, $f_V$, and $f_P$ denote
active force--length, force--velocity, and passive force--length curves.
Normalized muscle length $\ell_i^{(j)}$ and velocity $\nu_i^{(j)}$ (in optimal
lengths per second) follow MuJoCo's inelastic-tendon, zero-pennation muscle
model.

These forces drive multibody motion under model-specific inertia, passive
forces, and contacts~\cite{seth2011opensim,mujocoComputation}:
\begin{equation}
\begin{aligned}
&\dot{\boldsymbol{q}}=\boldsymbol{N}(\boldsymbol{q})\boldsymbol{v},\\
&\boldsymbol{M}(\boldsymbol{q};\boldsymbol{\eta}_i)\dot{\boldsymbol{v}}
+\boldsymbol{b}(\boldsymbol{q},\boldsymbol{v};\boldsymbol{\eta}_i)
+\boldsymbol{g}(\boldsymbol{q};\boldsymbol{\eta}_i)\\
&\quad=\boldsymbol{R}(\boldsymbol{q};\boldsymbol{\eta}_i)
\boldsymbol{F}(\boldsymbol{q},\boldsymbol{v},\boldsymbol{a};\boldsymbol{\eta}_i)
+\boldsymbol{\tau}_{\mathrm{pass}}(\boldsymbol{q},\boldsymbol{v};\boldsymbol{\eta}_i)\\
&\qquad+\boldsymbol{\tau}_{\mathrm{ext}}
+\boldsymbol{J}(\boldsymbol{q};\boldsymbol{\eta}_i)^\top\boldsymbol{\lambda}.
\end{aligned}
\label{eq:musculoskeletal-dynamics}
\end{equation}
Here, $\boldsymbol{M}$ is the generalized inertia matrix, while
$\boldsymbol{b}$ and $\boldsymbol{g}$ represent Coriolis/centrifugal and
gravitational forces. The matrix $\boldsymbol{R}$ maps muscle tensions to
generalized forces, and $\boldsymbol{N}$ maps generalized velocities to
coordinate derivatives. Non-muscle passive forces and external loads are
denoted by $\boldsymbol{\tau}_{\mathrm{pass}}$ and
$\boldsymbol{\tau}_{\mathrm{ext}}$ (passive muscle force enters
$\boldsymbol{F}$), while $\boldsymbol{J}^\top\boldsymbol{\lambda}$ accounts for
contact and friction forces.

Within this dynamics formulation, we learn a generalist policy with shared
parameters across individual models without fine-tuning:
\begin{equation}
\begin{aligned}
\text{Unconditioned:}\qquad\boldsymbol{e}_t
&\sim\pi_{\boldsymbol{\theta}_{\mathrm{U}}}(\,\cdot\mid\boldsymbol{o}_t),\\
\text{Conditioned:}\qquad\boldsymbol{e}_t
&\sim\pi_{\boldsymbol{\theta}_{\mathrm{C}}}(\,\cdot\mid\boldsymbol{o}_t,\boldsymbol{c}_i).
\end{aligned}
\label{eq:model-conditioned-control}
\end{equation}
Here, $\boldsymbol{o}_t$ contains musculoskeletal states, motion progress,
reference targets, and joint tracking errors, while
$\boldsymbol{c}_i=h(\boldsymbol{\eta}_i)$ summarizes the compiled model's
skeletal dimensions and total mass. The unconditioned form takes only $\boldsymbol{o}_t$, whereas
the conditioned form additionally conditions on $\boldsymbol{c}_i$. All trained
policies retain fixed parameters during evaluation.

\begin{figure}[t]
  \centering
  \includegraphics[width=88.6mm]{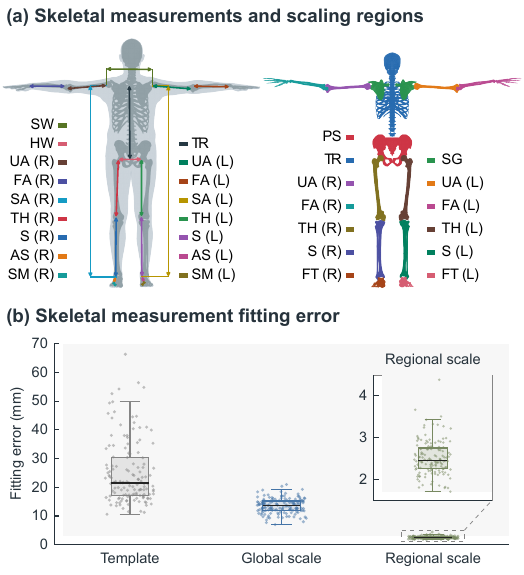}
  \caption{\textbf{Skeletal measurements and scaling regions for individual model construction.}
  (a) The 17 SKEL target skeletal measurements (left) and 13 MyoFullBody scaling regions (right). SW, shoulder width; HW, hip width; UA, upper arm; FA, forearm; TR, trunk; TH, thigh; S, shank; FT, foot; SA, shoulder--ankle vertical separation; AS, ankle--subtalar distance; SM, subtalar--metatarsophalangeal distance; PS, pelvis--sacrum; SG, shoulder girdle; R, right; L, left.
  (b) Mean absolute fitting error across the 17 measurements for Template, Global scale, and Regional scale ($n=141$). Tukey box plots show medians and interquartile ranges; the inset enlarges Regional scale.}
  \label{fig:model-construction}
\end{figure}

\subsection{Individual model construction}
We constructed individual musculoskeletal models through skeletal-guided regional scaling, which fitted the generic MyoFullBody template~\cite{li2026musclemimic} to 17 target skeletal measurements, followed by mass calibration to match recorded body mass. For the SMPL body-shape modality, SKEL provided these measurements in a reference T-pose~\cite{keller2023skel}. We fitted 13 positive regional scale factors $s_r=\exp(\xi_r)$ of the template by solving
\begin{equation}
\begin{aligned}
\boldsymbol\xi^\star&=\arg\min_{\boldsymbol\xi}\;
\sum_{k=1}^{17}\left[\frac{d_k(\boldsymbol\xi)-\hat d_k}{\tau_k}\right]^2
+\lambda\|\boldsymbol\xi-\xi_0\boldsymbol1\|_2^2\\
&\quad+\sum_{(r,v)\in\mathcal S}\frac{(\xi_r-\xi_v)^2}{\sigma_{\mathrm{sym}}^2}
+\sum_{(r,v)\in\mathcal E}\frac{(\xi_r-\xi_v)^2}{\sigma_{\mathrm{adj}}^2},\\
&\text{s.t.}\;\log s_{\min}\leq\xi_r\leq\log s_{\max},\quad r=1,\ldots,13.
\end{aligned}
\label{eq:regional-scale-estimation}
\end{equation}
Here, $d_k(\boldsymbol\xi)$ and $\hat d_k$ are the scaled-model and target skeletal measurements, and $\tau_k$ is the measurement tolerance. The remaining terms regularize overall scale, bilateral symmetry, and adjacent-region scales; $\mathcal S$ and $\mathcal E$ contain the bilateral and adjacent region pairs. The prior center was $\xi_0=\frac12\sum_{\zeta\in\{\mathrm L,\mathrm R\}}\log(\hat d_{\mathrm{SA},\zeta}/d_{\mathrm{SA},\zeta}^{\mathrm{template}})$, where $\hat d_{\mathrm{SA},\zeta}$ and $d_{\mathrm{SA},\zeta}^{\mathrm{template}}$ are the target and template shoulder-to-ankle vertical separations on side $\zeta$. Bounded nonlinear least squares yielded the Regional scale models (Supplementary Table~S1). Fig.~\ref{fig:model-construction}a illustrates the target skeletal measurements and scaling regions.

The fitted regional scale factors updated geometry and physical properties while preserving skeletal connectivity and musculotendon organization. For rigid body $b$,
\begin{equation}
\begin{aligned}
\boldsymbol p^{(b)\prime}&=s^{(\operatorname{pa}(b))}\boldsymbol p^{(b)},\\
\boldsymbol y^{(b)\prime}&=s^{(b)}\boldsymbol y^{(b)},\\
m^{(b)\prime}&=\alpha m^{(b)}(s^{(b)})^3,\\
\boldsymbol I^{(b)\prime}&=\alpha\boldsymbol I^{(b)}(s^{(b)})^5.
\end{aligned}
\label{eq:body-scaling}
\end{equation}
For musculotendon unit $j$,
\begin{equation}
\begin{aligned}
\kappa^{(j)}&=\frac{L^{(j)\prime}(\boldsymbol q_{\mathrm{ref}}')}{L^{(j)}(\boldsymbol q_{\mathrm{ref}})},\\
[L_{\min}^{(j)\prime},L_{\max}^{(j)\prime}]&=\kappa^{(j)}[L_{\min}^{(j)},L_{\max}^{(j)}],\\
F_{\max}^{(j)\prime}&=(\kappa^{(j)})^2F_{\max}^{(j)}.
\end{aligned}
\label{eq:musculotendon-scaling}
\end{equation}
Here, primes denote scaled quantities, $s^{(b)}$ is body $b$'s regional scale factor, and $\operatorname{pa}(b)$ is its parent. Body-frame origins $\boldsymbol p^{(b)}$ are expressed in the parent frame, whereas local coordinates $\boldsymbol y^{(b)}$ (geometry, attachment sites, joint anchors, and center-of-mass offsets) lie in body $b$'s frame. Mass $m^{(b)}$ and principal inertia $\boldsymbol I^{(b)}$ follow geometric similarity with a common density factor $\alpha$ calibrating total mass to recorded body mass. For muscle $j$, $L^{(j)}$ is path length, $[L_{\min}^{(j)},L_{\max}^{(j)}]$ is the operating range, and $\boldsymbol q_{\mathrm{ref}}$ and $\boldsymbol q_{\mathrm{ref}}'$ are the unscaled and scaled reference poses. Under geometric similarity and constant specific tension, $F_{\max}^{(j)}$ scales with physiological cross-sectional area~\cite{holzbaur2005model}.

We compared Regional scale against Global scale (1 isotropic scale factor fitted to the same 17 measurements with identical physical updates and mass calibration; Supplementary Section~S1) and the unchanged MyoFullBody template (Template).

The same pipeline handled motion-capture and monocular-image inputs. For Gait120, MoSh++ estimated SMPL-H parameters (SMPL with articulated hands~\cite{romero2017embodiedhands}) from marker recordings~\cite{mahmood2019amass}. For monocular-image construction, HSMR estimated SKEL shape from 1 image per individual~\cite{xia2025hsmr}; we scaled skeletal measurements by the ratio of measured height to the vertical extent of the estimated SKEL skin template in the standard pose. Both pathways calibrated total mass to recorded mass.

\subsection{Generalist full-body muscle control policy}
Within MyoAcross, we implemented 3 policy variants: an unconditioned
multilayer perceptron (MLP), a concatenation-conditioned policy (Concat), and a
feature-wise linear modulation policy (FiLM)~\cite{perez2018film}. All variants used separate actor
and critic networks encoding musculoskeletal states, motion progress, current
and future reference targets, and joint tracking errors into concatenated
embeddings. The actor output a diagonal Gaussian over latent actions
(state-dependent mean, learned state-independent standard deviations), mapped
via elementwise sigmoid to 416 muscle excitations; evaluation used the
deterministic mean (Supplementary Table~S2).

For morphology conditioning, we extracted model-derived features from each
compiled model (original, virtual, or asymmetric):
\begin{equation}
\boldsymbol{c}_i=(d_{i,1},\ldots,d_{i,17},M_i)^\top\in\mathbb{R}^{18},
\label{eq:morphology-features}
\end{equation}
where $d_{i,1},\ldots,d_{i,17}$ are the 17 skeletal measurements from
\textit{Individual model construction} and $M_i$ is total mass, log-transformed
and standardized using training-bank statistics.

\textbf{MLP} omitted $\boldsymbol{c}_i$ as an unconditioned ablation.
\textbf{Concat} used separate actor and critic morphology encoders whose
outputs were concatenated with the observation embeddings. \textbf{FiLM} used a
shared morphology encoder producing $\boldsymbol{z}_i$, which modulated each
actor and critic residual block:
\begin{equation}
\widetilde{\boldsymbol{h}}_{\ell}
=\bigl(\boldsymbol{1}+\boldsymbol{\gamma}_{\ell}(\boldsymbol{z}_i)\bigr)
\odot\operatorname{LN}(\boldsymbol{h}_{\ell})
+\boldsymbol{b}_{\ell}(\boldsymbol{z}_i).
\label{eq:film-conditioning}
\end{equation}
Here, $\boldsymbol{\gamma}_{\ell}$ and $\boldsymbol{b}_{\ell}$ output
feature-wise scale offsets and biases, $\operatorname{LN}$ denotes layer
normalization, and $\odot$ denotes elementwise multiplication. Each block
applied 2 affine layers with an intervening sigmoid linear unit (SiLU)
activation and a residual skip connection.

MLP, Concat, and FiLM shared identical training banks, reference motions,
backbone width/depth, and interaction budgets. We also evaluated an MLP trained on the generic MyoFullBody template (T-MLP).

\begin{figure}[!t]
  \centering
  \includegraphics[width=88.6mm]{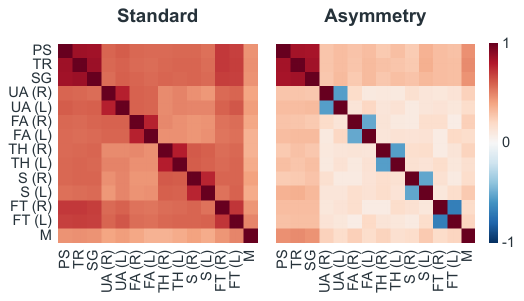}
  \caption{\textbf{Parameter correlations in virtual model augmentation.}
  Pearson correlations among relative changes (from original models) in regional scale factors and total mass ($M$) under standard (left) and asymmetry augmentation (right). Anatomical abbreviations follow Fig.~\ref{fig:model-construction}.}
  \label{fig:augmentation-correlations}
\end{figure}

\subsection{Virtual model augmentation}
We used virtual model augmentation to broaden the physical variation represented
in the training-model bank. Virtual models were generated around each original
training model through structured, correlated sampling of regional scale
factors and total mass:
\begin{equation}
\begin{aligned}
\widetilde{\boldsymbol{p}}_i
&=\boldsymbol{p}_i\odot(\boldsymbol{1}+\boldsymbol{\epsilon}),\\
\boldsymbol{\epsilon}
&\sim\mathcal{N}(\boldsymbol{0},\boldsymbol{\Sigma}),\qquad
\boldsymbol{\Sigma}=\boldsymbol{A}\boldsymbol{A}^{\top}.
\end{aligned}
\label{eq:virtual-model-sampling}
\end{equation}
Here, $\boldsymbol{p}_i=(s_{i,1},\ldots,s_{i,13},M_i)^\top$ contains the 13
regional scale factors and total mass of original model $i$. The loading matrix
$\boldsymbol{A}\in\mathbb{R}^{14\times18}$ combines 18 independent factors
representing shared size and limb variation, regional variation, bilateral
asymmetry, and residual mass variation. Fig.~\ref{fig:augmentation-correlations}
shows the resulting parameter correlations.

Proposals exceeding relative-change bounds were rejected; accepted parameters
generated virtual models via the same geometric, inertial, mass-calibration, and
musculotendon updates while retaining the original model's reference motions.
For Concat and FiLM, $\boldsymbol{c}_i$ was extracted from each compiled virtual
model, so augmentation altered simulated dynamics rather than only conditioning
inputs.

We evaluated 2 schemes: standard augmentation applied small bilateral
variations informed by tibial length differences in healthy
adults~\cite{vuurberg2022symmetry}, whereas asymmetry augmentation widened
bilateral and regional scale bounds based on clinical limb-length
discrepancies~\cite{riad2010discrepancy,kim2022discrepancy} (Supplementary
Section~S2 and Table~S3). V0 used only
original training models, whereas V10 added 10 virtual models per subject.
During training, subjects were sampled uniformly, followed by uniform sampling
across each subject's original and virtual models.

\subsection{Policy training}
We trained policies to maximize the expected discounted motion imitation reward
across sampled models and reference motions. The reward combined 8 body and
joint tracking terms~\cite{peng2018deepmimic} (Supplementary Table~S4):
\begin{equation}
R_t=\Delta t\sum_{h=1}^{8}w_h
\exp\!\left(-\frac{e_{t,h}}{\sigma_h^2}\right),
\label{eq:imitation-reward}
\end{equation}
where $e_{t,h}$ is the squared tracking error for term $h$ at control step $t$,
$w_h$ is its weight, $\sigma_h$ is its error scale, and $\Delta t$ is the
control time step. Body-pose targets were aligned with the simulated pelvis's
horizontal position and heading while preserving reference height; velocity
targets remained in world coordinates.

Parallel environment groups periodically received newly sampled models via this
subject--model scheme, with environments in each group sharing a model while
uniformly sampling reference motions. Episodes started from adaptively sampled
reference states favoring high-termination segments and ended upon completion,
time limit, or an aligned body-position error threshold.

Policies were optimized with proximal policy optimization
(PPO)~\cite{schulman2017ppo} in mjlab and MuJoCo
Warp~\cite{zakka2026mjlab,mujocoWarp} for 15,000 iterations (approximately
42\,h on 4 NVIDIA GeForce RTX 5090 GPUs) under identical budgets; the
template-trained MLP sampled only the MyoFullBody template. Full hyperparameters are provided in Supplementary Table~S5 and the released
codebase.

\subsection{Experimental protocol}
We used 141 subjects in the Archive of Motion Capture as Surface Shapes (AMASS)
format~\cite{mahmood2019amass}---55 from KIT~\cite{mandery2016unifying} and 86
from BMLmovi~\cite{ghorbani2021movi}---split into 110 training and 31 held-out
validation subjects. We selected 64 KIT source motions (8 clips in each of 8
categories: slow, medium-speed, and fast walking; backward walking; left and
right turns; and clockwise and counterclockwise circular walking) and
retargeted them to each original model using
GMR-Fit~\cite{li2026musclemimic,araujo2025retargeting}, denoting the retargeted
collection for model $i$ by $\mathcal{R}_i$. The monocular-image cohort comprised
88 MoVi subjects (63 training, 22 held-out, and 3 outside these splits) with
video-extracted images, measured height, and recorded body
mass~\cite{ghorbani2021movi}. For external biomechanical evaluation, Gait120
provided 120 male subjects with level-walking GRF and
sEMG recordings~\cite{boo2025gait120}, simulated using the 8
medium-speed KIT source motions retargeted to each model. All
participant data came from these public datasets under their respective license
terms (original ethics approvals: Queen's University for MoVi and KAIST IRB
KH2023-053 for Gait120); no new participants were recruited, and the author
pictured in Fig.~\ref{fig:framework}a provided informed consent for
publication.

For unilateral lower-limb shortening tests, thigh and shank regional scale
factors were reduced together on either side by 1\%--10\% (1-percentage-point
increments) with total mass recalibrated to its original value, updated
morphology features, and original reference motions retained. For input-sensitivity tests, physical models remained fixed. Morphology biases
(0\%--10\%, 1-percentage-point increments) were applied to $\boldsymbol{c}_i$
before log transformation and standardization. Reference-motion offsets
($0^\circ$--$10^\circ$, $1^\circ$ increments) perturbed bilateral hip, knee,
ankle, and subtalar targets in policy observations while leaving root motion,
joint velocities, initialization, reward, termination, and scoring unchanged.
Combined tests paired equal numerical magnitudes.

Control performance was quantified by completion (fraction of motions completed
without early termination) and tracking error (mean Euclidean position error
across 17 tracked rigid bodies in unaligned world coordinates, including
incomplete trials), averaged over motions within each subject and equally
across subjects (and over sign samples in perturbation tests). On Gait120,
biomechanical agreement was quantified by vertical GRF root-mean-square error
(RMSE, normalized by body weight, BW) over complete stance intervals ($\geq 0.2$\,s
at a 0.05\,BW threshold, 100 stance-phase points) and activation--sEMG
Pearson correlation over complete gait cycles (0.60--1.50\,s, 100 gait-phase
points, retaining dataset sEMG normalization), averaged across feet or channels
and subjects. Mean muscle activation and walking speed were averaged across
complete cycles, motions, and the 11 right-leg muscles corresponding to the sEMG
channels.

We computed 95\% percentile bootstrap confidence intervals (CIs) using 10,000
subject-level resamples (retaining all motions and perturbation samples within
each sampled subject), with paired subject resampling for FiLM versus MLP, V10
versus V0, perturbed versus matched clean baselines, and Gait120 model
conditions. Within Regional scale, associations between body mass and mean or
muscle-level activation were evaluated using Pearson correlations, partial
correlations, and ordinary least squares regressions (body-mass slopes per
10\,kg) adjusted for recorded height and simulated walking speed, with
covariate adjustment refitted in each of the 10,000 bootstrap resamples.

\section{Results}
\label{sec:results}

\subsection{Individual model construction accuracy}

Across the 141 AMASS subjects, skeletal-guided regional scaling (Regional scale)
achieved a mean absolute skeletal measurement fitting error of 2.51\,mm,
compared with 13.50\,mm for Global scale and 24.97\,mm for Template
(Fig.~\ref{fig:model-construction}b). This corresponds to reductions of
81.38\% relative to Global scale and 89.94\% relative to Template, preserving
individual skeletal proportions far more accurately than uniform scaling or the
generic template. All 141 models shared the same 416-muscle topology with
individual skeletal dimensions and calibrated mass; retargeting the 64 source
motions yielded 9,024 reference motions for the control evaluations
(Fig.~\ref{fig:framework}).

\subsection{Generalization across individual models}

\begin{figure}[!t]
  \centering
  \includegraphics[width=88.6mm]{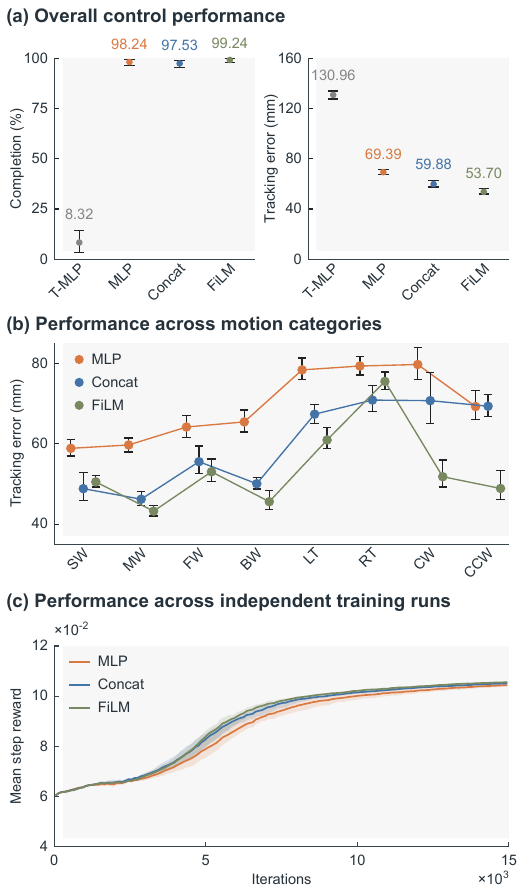}
  \caption{\textbf{Policy reuse across held-out individual models.}
  (a) Completion and tracking error for template-trained MLP (T-MLP) and MyoAcross variants (MLP, Concat, FiLM) on 31 held-out models without fine-tuning.
  (b) Tracking error across 8 locomotion categories (SW, slow; MW, medium; FW, fast; BW, backward walking; LT/RT, left/right turn; CW/CCW, clockwise/counterclockwise circling). Points show subject means with 95\% confidence intervals (CIs).
  (c) Mean step reward across independent training runs (100-iteration block averages; shaded bands show 95\% CIs across runs).}
  \label{fig:generalization}
\end{figure}

We evaluated cross-individual generalization on the 31 held-out individual
models (1,984 trials) without fine-tuning (Fig.~\ref{fig:generalization}a;
Supplementary Movie~1). Under standard augmentation (V10), FiLM within
MyoAcross achieved 99.24\% completion and a 53.70\,mm tracking error.

Comparison with the generic-model baseline highlighted the necessity of
multi-model training under individual morphological and inertial variations
(Fig.~\ref{fig:generalization}a; Supplementary Movie~2). Under the same
training budget, T-MLP achieved only 8.32\% completion
on the held-out models, whereas all 3 multi-model variants trained on the V10
bank achieved high completion: 98.24\% for MLP, 97.53\% for Concat, and
99.24\% for FiLM.

Within MyoAcross, explicit morphology conditioning substantially improved
tracking accuracy (Fig.~\ref{fig:generalization}a). FiLM reduced tracking error
to 53.70\,mm, compared with 69.39\,mm for MLP and 59.88\,mm for Concat
(relative reductions of 22.60\% and 10.33\%). FiLM outperformed both MLP and
Concat on all 31 held-out models, achieving lower error than MLP in all 8
locomotion categories and than Concat in 6, while Concat retained lower error
in slow walking and right turns (Fig.~\ref{fig:generalization}b). Together, multi-model training enables reliable completion across held-out
individuals, and FiLM conditioning consistently improves tracking accuracy
across individuals and most locomotion tasks.

Training curves across independent random seeds confirmed that these gains were
reproducible rather than sensitive to initialization
(Fig.~\ref{fig:generalization}c). All 3 variants progressed smoothly with
narrow 95\% CIs across runs. Beyond approximately 3,000
iterations, both morphology-conditioned policies (FiLM and Concat) consistently
achieved higher mean step rewards than the unconditioned MLP at matched
iterations, while FiLM maintained a slight advantage over Concat and converged
reliably across 15,000 iterations, mirroring its superior held-out tracking
accuracy.

\subsection{Virtual model augmentation and asymmetry transfer}

\begin{figure}[!t]
  \centering
  \includegraphics[width=88.6mm]{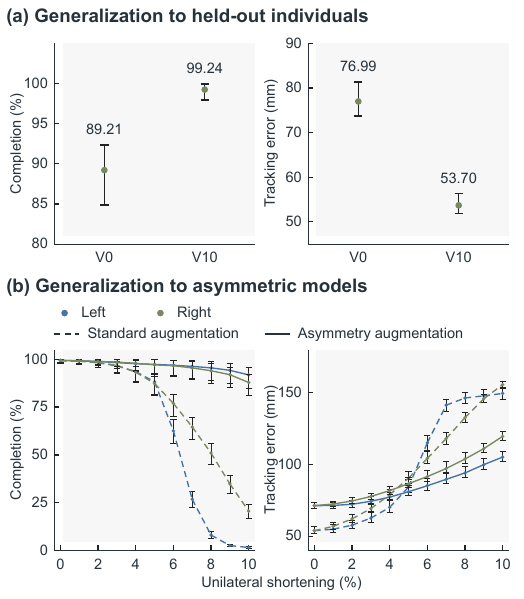}
  \caption{\textbf{Virtual model augmentation broadens generalization.}
  (a) Completion and tracking error of FiLM trained without augmentation (V0) versus 10 virtual models per subject (V10).
  (b) Transfer of standard versus asymmetry augmentation under 0\%--10\% unilateral thigh and shank shortening.
  Evaluated on 31 held-out models; points show subject means with 95\% CIs.}
  \label{fig:augmentation}
  \label{fig:asymmetry}
\end{figure}

To isolate virtual model augmentation under a fixed budget, we compared FiLM
trained on the 110 original models (V0) against FiLM trained with 10 virtual
models per subject (V10; 1,210 models) on the 31 held-out models
(Fig.~\ref{fig:augmentation}a). Augmentation increased completion from 89.21\%
to 99.24\%, a paired gain of 10.03 percentage points (95\% CI, 7.51--13.11
percentage points), and reduced tracking error from 76.99 to 53.70\,mm (a
30.25\% reduction). Completion improved in 30 of the 31 held-out
models and tracking error decreased in all 31, showing that structured
physical-parameter variation improves both stability and tracking accuracy
without additional training iterations.

We next evaluated whether tailoring the augmentation scheme extends policy
transfer to larger structural asymmetries using the unilateral lower-limb
shortening test models (1\%--10\% on either side; Fig.~\ref{fig:asymmetry}b).
Compared with standard augmentation, asymmetry augmentation (using FiLM with
V10) substantially improved transfer at larger shortening magnitudes. At 10\%
shortening, completion rose from 1.56\% to 91.83\% on the left side and from
20.56\% to 87.85\% on the right side, while tracking error fell from 149.48 to
105.14\,mm and from 155.60 to 119.60\,mm, respectively. On unshortened original
models (0\% shortening), completion remained high under both schemes (99.24\%
versus 99.45\%), whereas tracking error increased from 53.80 to 71.26\,mm.
Thus, broadening the augmentation distribution extends policy reuse to severe
unilateral lower-limb shortening, introducing a trade-off with tracking
accuracy on original models.

\subsection{Generalization to image-derived models and input sensitivity}

\begin{figure}[!t]
  \centering
  \includegraphics[width=88.6mm]{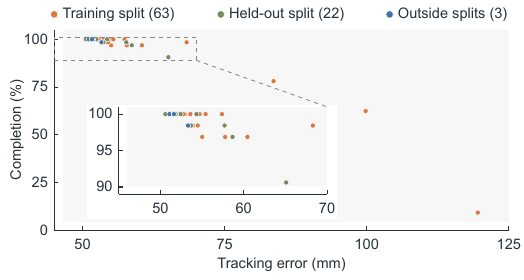}
  \caption{\textbf{Policy reuse on image-derived individual models.}
  Completion and tracking error across 88 image-derived models (training split, $n=63$; held-out, $n=22$; outside splits, $n=3$) without fine-tuning. Each point denotes a model; the inset enlarges the high-completion region.}
  \label{fig:image-entry}
\end{figure}

We next evaluated whether the pretrained FiLM policy (V10, standard
augmentation) transfers to models constructed from monocular-image shape
estimates paired with height calibration and recorded body mass, without
fine-tuning (Fig.~\ref{fig:framework}a; Fig.~\ref{fig:image-entry}). Across all
88 image-derived models, FiLM completed 5,514 of 5,632 motions (97.90\%
completion, 54.38\,mm tracking error); on the 22 held-out individuals, it
completed 1,398 of 1,408 motions (99.29\% completion, 52.89\,mm tracking
error; cohort breakdown and outlier diagnostics in Supplementary Section~S3 and
Table~S6). These results confirm that monocular-image construction provides a
low-cost entry point for cross-individual policy reuse, including on held-out
individuals.

\begin{figure}[!t]
  \centering
  \includegraphics[width=88.6mm]{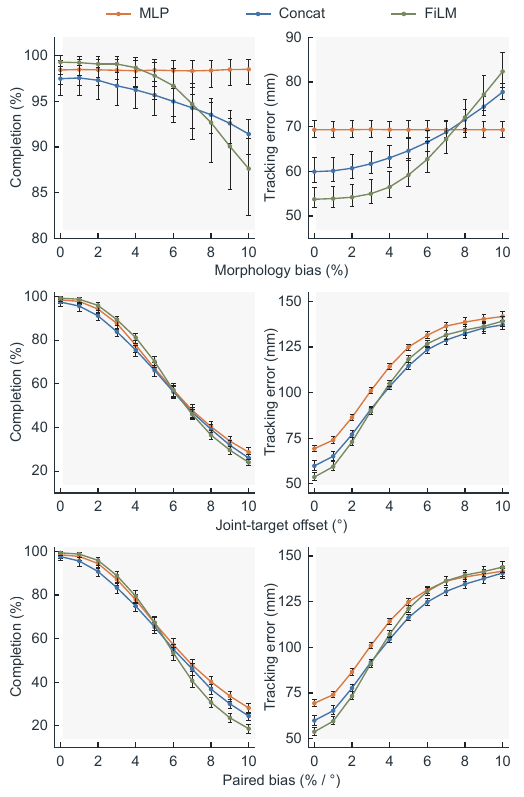}
  \caption{\textbf{Policy sensitivity to input perturbations.}
  Completion and tracking error on 31 held-out models under morphology feature biases (top), reference-motion joint-angle offsets (middle), and combined perturbations (bottom). MLP receives no morphology inputs. Points show subject means with 95\% CIs.}
  \label{fig:sensitivity}
\end{figure}

To separate generalization across physical model parameters from sensitivity to
inaccurate policy inputs, we perturbed morphology features and reference-motion
inputs separately and in combination while keeping physical models and policy
parameters fixed on the 31 held-out individual models
(Fig.~\ref{fig:sensitivity}; \mbox{Supplementary Movie~2}). Under morphology
feature perturbations (Fig.~\ref{fig:sensitivity}, top row), FiLM maintained
99.09\% and 97.82\% completion at 2\% and 5\% biases, respectively. At 5\% bias, tracking error increased from 53.74\,mm (unperturbed baseline) to
59.18\,mm, remaining lower than Concat (64.62\,mm). At 10\% bias, completion declined to 87.62\% for FiLM and 91.41\%
for Concat, whereas MLP was unaffected because it receives no morphology
inputs; FiLM's greater sensitivity at large biases alongside its lower
unperturbed tracking error reflects its stronger reliance on morphology
conditioning.

Reference-motion input offsets degraded stability more sharply than morphology
biases (Fig.~\ref{fig:sensitivity}, middle row): FiLM completion dropped to
95.92\%, 70.25\%, and 24.01\% at joint-angle offsets of $2^\circ$, $5^\circ$,
and $10^\circ$, respectively. Under combined perturbations (Fig.~\ref{fig:sensitivity}, bottom row), FiLM
maintained 95.82\% completion at 2\% morphology bias paired with a $2^\circ$
reference offset, but decreased to 18.70\% completion at 10\% bias paired with
a $10^\circ$ offset (compared with 28.19\% completion for MLP and 24.50\% for
Concat). Together, these evaluations show that policy reuse transfers reliably to
image-derived models and tolerates modest input errors, while remaining
sensitive to large input perturbations---particularly in reference-motion
kinematics.

\subsection{Biomechanical validation}
\label{sec:biomechanical-validation}

\begin{figure}[!t]
  \centering
  \includegraphics[width=88.6mm]{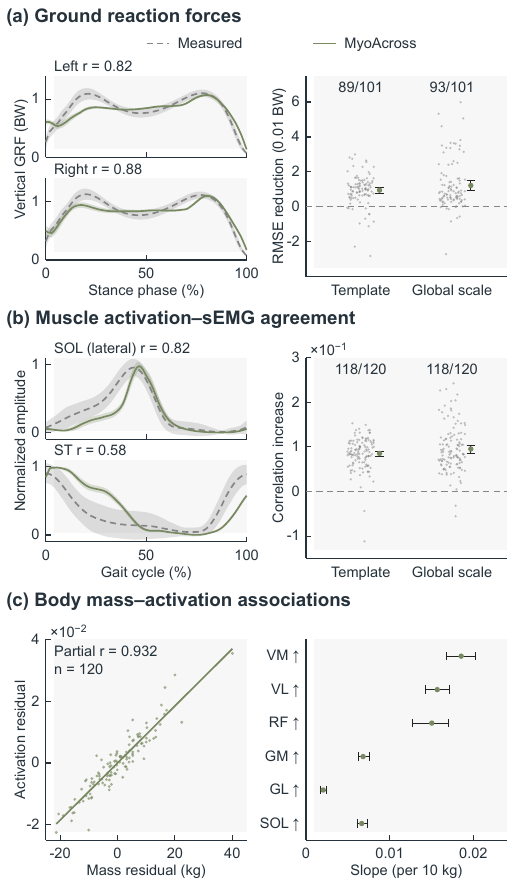}
  \caption{\textbf{Biomechanical validation across Gait120 subjects.}
  MyoAcross denotes the pretrained FiLM policy controlling Regional scale models.
  (a) Vertical GRF normalized by body weight (BW, left) and RMSE reductions relative to Template and Global scale (right).
  (b) Normalized muscle activation and sEMG profiles for soleus (SOL) and semitendinosus (ST) (left), and activation--sEMG correlation gains (right).
  (c) Mass--activation partial-regression residuals (left) and muscle-specific slopes (right), adjusted for height and walking speed; arrows indicate experimental loading trends~\cite{damewood2025generation,mcgowan2008independent}.
  Curves show means $\pm1$ standard deviation (SD); $r$, within-subject Pearson correlation; error bars, 95\% CIs; VM/VL, vastus medialis/lateralis; RF, rectus femoris; GM/GL, gastrocnemius medialis/lateralis.}
  \label{fig:biomechanical-validation}
\end{figure}

To evaluate whether the framework supports biomechanical comparisons on an
independent human dataset, we compared Template, Global scale, and Regional
scale models across the 120 Gait120 subjects~\cite{boo2025gait120} using the
same pretrained FiLM policy (V10, standard augmentation) with fixed policy
parameters across 8 medium-speed KIT source motions retargeted to each
model~\cite{mandery2016unifying} (Fig.~\ref{fig:framework}a;
Fig.~\ref{fig:biomechanical-validation}). Completion reached 100\% for both
Template and Regional scale models and 95.83\% for Global scale models.

Regional scale models improved agreement between simulated and measured vertical
GRFs (Fig.~\ref{fig:biomechanical-validation}a). Across
the 101 subjects with valid bilateral GRF measurements, mean subject-level
RMSE normalized by BW was 0.139\,BW for
Regional scale, compared with 0.149\,BW for Template and 0.151\,BW for Global
scale. Paired RMSE reductions were 0.0095\,BW relative to Template (95\%
CI, 0.0076--0.0112\,BW) and 0.0122\,BW relative to
Global scale (95\% CI, 0.0095--0.0150\,BW).

Regional scale models also improved agreement between simulated muscle
activations and measured sEMG
(Fig.~\ref{fig:biomechanical-validation}b; Supplementary Movie~3). Across all
120 subjects, mean within-subject Pearson correlations averaged across 12 sEMG
channels reached 0.407 for Regional scale, compared with 0.321 for Template and
0.311 for Global scale, improving in 118 of 120 subjects relative to both
Template and Global scale. Restricting the Regional scale versus Global scale
comparison to identical completed motions within each subject yielded consistent
gains: vertical GRF RMSE decreased by 0.0123\,BW (95\% CI,
0.0097--0.0151\,BW) and activation--sEMG correlation increased by 0.0952 (95\%
CI, 0.0867--0.1041). These improvements confirm the
biomechanical value of skeletal-guided regional scaling when evaluated with
fixed policy parameters and common source motions (policy-variant comparisons
on Gait120 are summarized in Supplementary Table~S7).

Within the Regional scale models, mean cycle-averaged muscle activation across
the 11 right-leg muscles corresponding to the sEMG channels increased strongly
with body mass across the 120 subjects (Pearson $r=0.823$), remaining high
after adjustment for recorded height and simulated walking speed (partial
$r=0.932$; 95\% CI, 0.902--0.955;
Fig.~\ref{fig:biomechanical-validation}c). This adjusted association remained
positive across all 8 source motions (partial $r=0.875$--0.950). At the
individual muscle level (Fig.~\ref{fig:biomechanical-validation}c; full
11-muscle statistics in Supplementary Table~S8b), vastus
medialis, vastus lateralis, and rectus femoris exhibited the largest positive
regression slopes among the analyzed muscles, alongside positive associations in
gastrocnemius medialis, gastrocnemius lateralis, and soleus. These directions
align with experimental human loading studies reporting increased quadriceps,
gastrocnemius, and soleus EMG during loaded
walking~\cite{damewood2025generation,mcgowan2008independent}.
Together, these findings show that a generalist policy supports biomechanical
comparisons across individual models and reproduces physiologically plausible,
body-mass-associated activation patterns.

\section{Discussion}
\label{sec:discussion}

MyoAcross bridges skeletal-guided individual model construction with a
generalist morphology-conditioned policy, enabling scalable individualized
musculoskeletal simulation across diverse individuals without fine-tuning.
Whereas prior modeling and calibration workflows lacked a generalist full-body
policy~\cite{werling2023addbiomechanics,berman2024personalizing}, and earlier
cross-morphology muscle control studies focused on simplified planar
arms~\cite{crowder2022learning} or omitted external biomechanical
evaluation~\cite{lee2019scalable,feng2023musclevae,kim2025freemusco,an2025arnold},
MyoAcross unifies accessible model construction, a generalist policy, and
external validation in one framework. The sharp gap between the template-trained
MLP (8.32\% completion) and MyoAcross policies (up to 99.24\%) confirms that
individual skeletal and mass variations substantially alter dynamics,
necessitating multi-model training and morphology-aware control.

Our ablations show that morphology conditioning and virtual model augmentation
address complementary needs for cross-individual reuse: runtime awareness of
active model properties and training-time exposure to physical variation. Under
the same interaction budget, FiLM consistently outperformed MLP and Concat in
tracking accuracy, while adding virtual models (V0 to V10) improved both
completion and tracking accuracy. The unilateral shortening tests further show
that tailoring the augmentation distribution extends policy transfer to
pronounced structural asymmetries motivated by clinical limb-length
discrepancies~\cite{riad2010discrepancy,kim2022discrepancy}. Expanding training
coverage toward severe asymmetries, however, degraded tracking accuracy on
unshortened models, reflecting finite policy capacity and interaction budget
shared across a broader morphological distribution.

Generalization to image-derived models extends MyoAcross beyond SMPL
body shapes to an accessible monocular-image entry point. Alongside the
controlled perturbation tests, these results clarify upstream input
requirements. Whereas image-derived construction alters the physical model
itself, fixed-model perturbations revealed an asymmetric sensitivity profile:
FiLM tolerated modest morphology feature biases far better than kinematic
reference-motion offsets, which sharply degraded completion at larger angles.
Moreover, FiLM's higher sensitivity at large morphology biases---paired with its
superior unperturbed accuracy---confirms active reliance on morphology features
rather than morphology-invariant heuristics. Because policy inputs depend on
retargeted trajectories, these findings impose stricter accuracy requirements on
source motions and retargeting while demonstrating practical tolerance to
morphological estimation errors.

The external Gait120 evaluation confirms that preserving individual skeletal
proportions yields measurable biomechanical gains beyond kinematic tracking.
Because Template, Global scale, and Regional scale models were controlled by the
same pretrained FiLM policy with fixed parameters and identical source motions,
the reductions in vertical GRF RMSE and gains in activation--sEMG correlation
directly isolate the contribution of skeletal-guided regional scaling. Within
Regional scale models, mean activation increased strongly with body mass even
after adjusting for height and walking speed (partial $r=0.932$), with positive
slopes in the quadriceps and plantar flexors. Although this analysis reflects
between-individual variation rather than within-person added-load
trials~\cite{damewood2025generation,mcgowan2008independent}, its directional
agreement supports the physiological plausibility of simulated recruitment
patterns.

Several limitations define the current scope. First, all models share a
416-muscle topology with inelastic tendons, zero pennation, and geometrically
scaled maximum isometric forces; extending the framework to capture
neuromuscular impairments, spasticity, altered tendon compliance, and
pathological joint mechanics requires further investigation. Second, the current framework uses a fixed set of
64 locomotion source motions rather than arbitrary non-locomotion behaviors.
Because retargeting scales with the product of source motions and individual
models, expanding motion libraries across large model banks incurs high
computational cost; accelerating cross-individual retargeting will be key to
broader training motion repertoires. Third, image-derived model construction still
requires measured height for calibration, as monocular mesh recovery methods
such as HSMR~\cite{xia2025hsmr} preserve relative proportions well but do not
reliably recover absolute metric scale.

Despite these boundaries, coupling accessible individual model construction with
a generalist policy offers a scalable basis for systematic cross-individual
musculoskeletal simulation. In rehabilitation engineering and wearable robotics,
musculoskeletal simulation with DRL has supported prosthetic gait
modeling~\cite{devree2021deep,ogum2024walking}, rehabilitation robot trajectory
generation~\cite{khan2025reinforcement}, and exoskeleton assistance policy
deployment~\cite{luo2024exoskeleton}. Integrating MyoAcross models with wearable
device dynamics could enable scalable cross-individual evaluation of assistive
strategies and DRL-based exoskeleton assistance policy design without training a
separate policy for every user.

\section{Conclusion}
\label{sec:conclusion}

MyoAcross provides a framework for scalable individualized musculoskeletal
simulation that connects skeletal-guided individual model construction with a
generalist morphology-conditioned policy. Combining model-derived morphology
conditioning with virtual model augmentation enables full-body reference-motion
tracking across held-out individual models without fine-tuning. Evaluations
across SMPL body shape, monocular image, and external Gait120 marker-based motion capture pathways
demonstrate policy reuse across input modalities and confirm the biomechanical
value of preserving individual skeletal proportions. Within its evaluated
musculoskeletal topology and locomotion repertoire, MyoAcross offers a practical
foundation for cross-individual biomechanical analysis, rehabilitation
engineering, and exoskeleton assistance policy design.

\section*{Data and Code Availability}
Third-party datasets are publicly available from AMASS~\cite{mahmood2019amass,mandery2016unifying,ghorbani2021movi}, MoVi~\cite{ghorbani2021movi}, and Gait120~\cite{boo2025gait120}. Code and pretrained checkpoints are available on \href{https://github.com/woodater-thu/MyoAcross}{GitHub} and \href{https://huggingface.co/woodater/myoacross-checkpoints}{Hugging Face}, respectively, with license terms detailed on the respective repository pages.

\bibliographystyle{IEEEtran}
\bibliography{references}

\newgeometry{margin=0.75in}
\pagestyle{plain}
\setcounter{table}{0}
\setcounter{figure}{0}
\setcounter{equation}{0}
\setcounter{subsection}{0}
\renewcommand{\figurename}{Supplementary Figure}
\renewcommand{\tablename}{Supplementary Table}
\renewcommand{\refname}{Supplementary References}
\renewcommand{\thefigure}{S\arabic{figure}}
\renewcommand{\thetable}{S\arabic{table}}
\renewcommand{\theequation}{S\arabic{equation}}
\renewcommand{\theHfigure}{supp.\arabic{figure}}
\renewcommand{\theHtable}{supp.\arabic{table}}
\renewcommand{\theHequation}{supp.\arabic{equation}}
\renewcommand{\theHsubsection}{supp.\arabic{subsection}}
\renewcommand{\thesubsection}{Supplementary Section S\arabic{subsection}:}
\setlength{\parindent}{0pt}
\setlength{\parskip}{5pt}
\setlength{\floatsep}{6pt plus 1pt minus 1pt}
\setlength{\textfloatsep}{6pt plus 1pt minus 1pt}
\setlength{\intextsep}{6pt plus 1pt minus 1pt}
\makeatletter
\renewcommand{\@seccntformat}[1]{\csname the#1\endcsname\ }
\setlength{\@fptop}{0pt}
\setlength{\@fpsep}{6pt}
\setlength{\@fpbot}{0pt plus 1fil}
\makeatother

\raggedbottom
\begin{center}
{\fontsize{20}{24}\selectfont\bfseries Supplementary Material\par}
\vspace{12pt}
{\fontsize{16}{20}\selectfont\bfseries MyoAcross: A Framework for Scalable Individualized Musculoskeletal\\
Simulation with a Generalist Morphology-Conditioned Policy\par}
\end{center}

\section*{Supplementary Sections}

\subsection{Global Scale and Template Comparators}
\label{sec:supp-global-fitting}
Global scale models were constructed by fitting 1 isotropic scale factor $s=\exp(\xi)$ to the same 17 SKEL skeletal targets used for Regional scale, followed by identical physical-parameter updates and mass calibration~\cite{clemente2024predictive}:
\begin{equation}
\xi^\star=\underset{\log s_{\min}\leq\xi\leq\log s_{\max}}{\arg\min}
\left\{\sum_{k=1}^{17}\left[\frac{d_k^{\mathrm{global}}(\xi)-\hat d_k}{\tau_k}\right]^2+\lambda_{\mathrm g}(\xi-\xi_0)^2\right\}.
\label{eq:supp-global-fit}
\end{equation}
Here, $d_k^{\mathrm{global}}(\xi)$ is the $k$th uniformly scaled measurement, $\hat d_k$ is its target, and $\lambda_{\mathrm g}$ weights the overall-scale prior. Fitting used the same 17 targets, tolerances $\tau_k$, scale bounds, and prior center $\xi_0$ as regional scaling (Supplementary Table~\ref{tab:supp-fitting-parameters}), scaling all bodies uniformly except the global root. Template retained the original MyoFullBody geometry and physical parameters without scaling or mass calibration.

\subsection{Rationale for Bilateral Asymmetry Sampling}
\label{sec:supp-asymmetry}
Bilateral length asymmetry was quantified as
\begin{equation}
D=\frac{2|L_R-L_L|}{L_R+L_L},
\qquad
\mathbb E[D]\simeq 2\sigma_a\sqrt{2/\pi},
\label{eq:supp-asymmetry-scale}
\end{equation}
where $L_R$ and $L_L$ are the right and left segment lengths. Assuming an initially symmetric pair perturbed by an opposite-sign factor $a\sim\mathcal N(0,\sigma_a^2)$, the 0.77\% mean relative tibial length difference reported in healthy adults~\cite{vuurberg2022symmetry} yields $\sigma_a\simeq 0.00483$, motivating the standard-augmentation prior (Supplementary Table~\ref{tab:supp-priors}).

Clinical reports of larger limb-length discrepancies in hemiplegic cerebral palsy motivated the wider asymmetry-augmentation scale~\cite{riad2010discrepancy,kim2022discrepancy}. Original-model asymmetry, combined sampling factors, and rejection bounds jointly determine the resulting training distributions.

\subsection{Failure-Case and Outlier Diagnostics}
\label{sec:supp-diagnostics}
\textbf{Concentration of incomplete trials.} Incomplete trials were concentrated in a few models. Across the 31 held-out Archive of Motion Capture as Surface Shapes (AMASS) individuals, MoVi Subject 44 accounted for 12 of 15 incomplete trials under feature-wise linear modulation (FiLM; V10, standard augmentation). Across the 88 image-derived models, Subjects 48, 75, and 80 (all in the original training split; Supplementary Table~\ref{tab:supp-image-subsets}) accounted for 96 of 118 incomplete trials. Under 10\% unilateral lower-limb shortening, asymmetry augmentation achieved at least 80\% completion in 30 of 31 models for left-sided shortening and 29 of 31 for right-sided shortening, with Subject 44 being the only model below 50\% in both conditions.

\textbf{Physical differences in low-completion image-derived models.} Comparing original and image-derived models for Subjects 48, 75, and 80 under the same FiLM checkpoint showed completion dropping from 95.31\%--100\% (original) to 12.50\%--79.69\% (image-derived). At identical recorded mass, these image-derived models had 14.57\%--22.19\% narrower hip widths, yielding 11.69\%--23.87\% lower summed maximum isometric muscle force relative to body weight because force scales with geometric area while total mass is calibrated separately. Reduced relative force capacity and altered retargeted kinematics therefore account for these outliers. Conversely, image-derived construction did not uniformly degrade control: Subject 44 reached 90.63\% completion on its image-derived model versus 81.25\% on its original model.

\clearpage
\section*{Supplementary Tables}

\begin{table}[H]
\centering
\caption{Scale-fitting parameters for Regional scale and Global scale.}
\label{tab:supp-fitting-parameters}
\small
\begin{tabular*}{\textwidth}{@{\extracolsep{\fill}}p{0.58\textwidth}p{0.38\textwidth}@{}}
\toprule
Parameter & Value \\
\midrule
Scale bounds, $s_{\min},s_{\max}$ & 0.7, 1.3 \\
Regional overall-scale prior weight, $\lambda$ & 1 \\
Global scale prior weight, $\lambda_{\mathrm g}$ & 1 \\
Bilateral regularization scale, $\sigma_{\mathrm{sym}}$ & 0.03 \\
Adjacency regularization scale, $\sigma_{\mathrm{adj}}$ & 0.15 \\
Hip-width tolerance & 5\,mm \\
Trunk-length tolerance & 7.5\,mm \\
Ankle-to-subtalar distance tolerance & 15\,mm \\
Remaining measurement tolerances & 10\,mm \\
Initial log-scale & $\xi_0$, clipped to bounds \\
Maximum function evaluations & 100 \\
Regional scale acceptance & Solver success and every absolute normalized measurement error $\leq 4$ \\
Global scale acceptance & Solver success \\
\bottomrule
\end{tabular*}
\par\smallskip\raggedright\footnotesize
Bilateral pairs comprise the upper arms, forearms, thighs, shanks, and feet. Adjacent pairs comprise pelvis--sacrum and trunk; trunk and shoulder girdle; and, on each side, shoulder girdle and upper arm, upper arm and forearm, pelvis--sacrum and thigh, thigh and shank, and shank and foot.
\end{table}

\begin{table}[H]
\centering
\caption{Policy observations and network configuration.}
\label{tab:supp-network}
\small
\begin{tabular*}{\textwidth}{@{}p{0.16\textwidth}@{\extracolsep{\fill}}p{0.11\textwidth}p{0.42\textwidth}p{0.23\textwidth}@{}}
\toprule
\multicolumn{4}{@{}l}{\textbf{a. Observation groups and encoders}} \\
\addlinespace[3pt]
Group & Dimension & Contents & Encoder widths \\
\midrule
Root & 9 & Gravity direction and linear/angular velocity in the pelvis frame & 32, 64 \\
Joint & 732 & Positions, velocities, reference values, and differences for 122 scalar joints & 1,024, 1,024 \\
Muscle & 2,080 & Length, velocity, force, excitation, and activation for 416 muscles & 2,048, 2,048 \\
Reference & 10 & Current pelvis target position, 6D orientation, and normalized motion progress & 32, 64 \\
Contact & 4 & Right/left foot and toe touch forces & 16, 32 \\
Future reference & $4\times 67$ & Positions of global root and 17 tracking bodies, pelvis orientation and velocities, and time offset in the current pelvis frame & 128, 128 per frame; 256 after ordered concatenation \\
Morphology & 18 & 17 skeletal measurements and body mass & 64, 128, 256 \\
\bottomrule
\end{tabular*}
\par\medskip
\begin{tabular*}{\textwidth}{@{}p{0.32\textwidth}@{\extracolsep{\fill}}p{0.64\textwidth}@{}}
\toprule
\multicolumn{2}{@{}l}{\textbf{b. Network configuration}} \\
\addlinespace[3pt]
Component & Setting \\
\midrule
Residual backbone & Width 2,048; 6 blocks; SiLU activation; residual scale 0.5 \\
Actor and critic heads & Hidden widths 1,024, 512 for each head \\
FiLM generator & Hidden width 256; layer normalization without affine parameters \\
Future-reference encoding & Shared encoder across frames within each network; time offsets 0.1, 0.2, 0.3, 0.4\,s \\
Observation normalization & Separate actor and critic statistics; minimum standard deviation 0.001 \\
Morphology normalization & Log features in meters and kilograms; training-bank mean and population standard deviation, equal weight per model \\
Latent-action initialization & Standard deviation 0.5; mean-head bias $\log[0.1/(1-0.1)]$ \\
\bottomrule
\end{tabular*}
\par\smallskip\raggedright\footnotesize
Encoder widths include the output layer. SiLU, sigmoid linear unit.
\end{table}

\begin{table}[H]
\centering
\caption{Virtual model sampling parameters for standard and asymmetry augmentation.}
\label{tab:supp-priors}
\small
\begin{tabular*}{\textwidth}{@{\extracolsep{\fill}}lrr@{}}
\toprule
Parameter & \shortstack{Standard\\augmentation} & \shortstack{Asymmetry\\augmentation} \\
\midrule
Global size factor SD & 0.010 & 0.010 \\
Axial factor SD & 0.005 & 0.005 \\
Upper-limb factor SD & 0.005 & 0.005 \\
Lower-limb factor SD & 0.005 & 0.005 \\
Upper arm, forearm, thigh, shank regional SD & 0.008 & 0.008 \\
Pelvis, trunk, shoulder girdle, foot regional SD & 0.005 & 0.005 \\
Paired-region asymmetry factor SD & 0.005 & 0.025 \\
Mass--size coefficient & 2.0 & 2.0 \\
Mass residual SD & 0.030 & 0.030 \\
Absolute relative regional scale bound & 0.06 & 0.12 \\
Absolute relative mass bound & 0.15 & 0.15 \\
\bottomrule
\end{tabular*}
\par\smallskip\raggedright\footnotesize
All parameters are dimensionless. SD, standard deviation.
\end{table}

\begin{table}[H]
\centering
\caption{Motion imitation reward terms and parameters.}
\label{tab:supp-reward}
\small
\begin{tabular*}{\textwidth}{@{\extracolsep{\fill}}llrl@{}}
\toprule
Tracking term & Error $e_{t,h}$ & $w_h$ & $\sigma_h$ \\
\midrule
Pelvis position & Squared distance & 0.5 & 0.3\,m \\
Pelvis orientation & Squared angle & 0.5 & 0.4\,rad \\
Body position & Mean squared distance & 2.0 & 0.3\,m \\
Body orientation & Mean squared angle & 0.5 & 0.4\,rad \\
Body linear velocity & Mean squared vector error & 0.8 & 1.0\,m/s \\
Body angular velocity & Mean squared vector error & 0.5 & 3.14\,rad/s \\
Joint position & Mean squared scalar error & 0.6 & 0.3\,rad \\
Joint velocity & Mean squared scalar error & 0.2 & 2.0\,rad/s \\
\bottomrule
\end{tabular*}
\par\smallskip\raggedright\footnotesize
Body and joint errors were averaged over 17 rigid bodies (pelvis, torso, head, and bilateral humerus, ulna, lunate, femur, tibia, calcaneus, and toes) and 15 hinge joints (trunk flexion--extension, lateral bending, and axial rotation; bilateral hip flexion, adduction, and rotation; knee flexion; ankle; and subtalar angles).
\end{table}

\begin{table}[H]
\centering
\caption{Simulation and policy-training parameters.}
\label{tab:supp-training}
\small
\begin{tabular*}{\textwidth}{@{\extracolsep{\fill}}p{0.54\textwidth}p{0.42\textwidth}@{}}
\toprule
Parameter & Value \\
\midrule
\multicolumn{2}{@{}l}{\textbf{a. Simulation and termination}} \\
Simulation time step & 0.002\,s \\
Control time step & 0.02\,s \\
Simulation backend & mjlab 1.6.0 and MuJoCo Warp \\
Episode time limit & 10\,s \\
Pelvis position / mean aligned body-position error limit & $0.3\times$ active model height \\
Pelvis orientation error limit & 1.0\,rad \\
\midrule
\multicolumn{2}{@{}l}{\textbf{b. Training and reference-state sampling}} \\
GPUs per run & 4 (NVIDIA GeForce RTX 5090) \\
Training iterations & 15,000 \\
Environments per GPU & 2,048 \\
Rollout steps per iteration & 32 \\
Adaptive sampling bins & $B=\lfloor L/f\rfloor+1$ \\
Termination-count moving-average weight & 0.001 per control step \\
Additive sampling weight per bin & $0.1/B$ \\
\midrule
\multicolumn{2}{@{}l}{\textbf{c. Optimization}} \\
Optimizer & AdamW \\
Weight decay & 0 \\
Initial learning rate & $3\times 10^{-5}$ \\
Learning-rate bounds & $10^{-5}$ to $10^{-2}$ \\
Target KL divergence & 0.01 \\
Learning-rate adjustment factor & 1.5 \\
KL thresholds for learning-rate decrease / increase & $>2\times$ target / $0<\mathrm{KL}<0.5\times$ target \\
Discount factor & 0.99 \\
Generalized advantage estimation parameter & 0.95 \\
PPO clipping parameter & 0.2 \\
Learning epochs per training iteration & 3 \\
Minibatches per learning epoch & 4 \\
Value loss coefficient & 1.0 \\
Entropy coefficient & 0 \\
Maximum gradient norm & 1.0 \\
\bottomrule
\end{tabular*}
\par\smallskip\raggedright\footnotesize
KL, Kullback--Leibler divergence; PPO, proximal policy optimization; $L$, reference-motion frame count; $f$, frame rate (Hz). Adaptive sampling weights equal running episode-termination counts per bin plus the additive weight, normalized to sum to 1. Initial frames are sampled uniformly within the selected bin, excluding the final frame.
\end{table}

\begin{table}[H]
\centering
\caption{Image-derived model performance by cohort membership.}
\label{tab:supp-image-subsets}
\small
\begin{tabular*}{\textwidth}{@{\extracolsep{\fill}}lrrrr@{}}
\toprule
Original cohort membership & Models & Trials & Completion (\%) & Tracking error (mm) \\
\midrule
Full cohort & 88 & 5,632 & 97.90 & 54.38 \\
Training & 63 & 4,032 & 97.35 & 55.01 \\
Held-out & 22 & 1,408 & 99.29 & 52.89 \\
Outside original cohort & 3 & 192 & 99.48 & 52.01 \\
\bottomrule
\end{tabular*}
\par\smallskip\raggedright\footnotesize
Evaluated with the pretrained FiLM policy (V10, standard augmentation) without fine-tuning.
\end{table}

\begin{table}[H]
\centering
\caption{Control and biomechanical performance across policy variants on Gait120 Regional scale models.}
\label{tab:supp-gait120-results}
\small
\begin{tabular*}{\textwidth}{@{\extracolsep{\fill}}lrrrr@{}}
\toprule
Policy & Completion (\%) & Tracking error (mm) & GRF RMSE (BW) & sEMG $r$ \\
\midrule
FiLM (V10, standard) & 100 & 45.05 & 0.139 & 0.406 \\
Concat (V10, standard) & 100 & 50.08 & 0.150 & 0.424 \\
MLP (V10, standard) & 100 & 57.82 & 0.156 & 0.160 \\
FiLM (V0) & 99.90 & 53.50 & 0.151 & 0.105 \\
FiLM (V10, asymmetry) & 100 & 66.25 & 0.151 & 0.348 \\
\bottomrule
\end{tabular*}
\par\smallskip\raggedright\footnotesize
Subject means across 8 reference motions ($n=120$ for control and 12-channel surface electromyography, sEMG; $n=101$ for bilateral vertical ground reaction force, GRF). FiLM (V0) completed 959 of 960 motions; waveform metrics use completed motions. MLP, multilayer perceptron; Concat, concatenation-conditioned policy; RMSE, root-mean-square error; BW, body weight.
\end{table}

\begin{table}[H]
\centering
\caption{Muscle-level associations with body mass across the 11 right-leg Gait120 sEMG muscles.}
\label{tab:supp-muscle-mass-associations}
\small
\begin{tabular*}{\textwidth}{@{\extracolsep{\fill}}llll@{}}
\toprule
Muscle & Slope per 10\,kg [95\% CI] & Partial $r$ [95\% CI] & Adjusted $P$ \\
\midrule
\multicolumn{4}{@{}l}{\textbf{a. Global scale}} \\
Biceps femoris, long head & 0.0199 [0.0159, 0.0243] & 0.887 [0.869, 0.924] & $1.98\times 10^{-40}$ \\
Gastrocnemius lateralis & 0.0045 [0.0034, 0.0058] & 0.861 [0.838, 0.939] & $1.09\times 10^{-35}$ \\
Gastrocnemius medialis & 0.0088 [0.0080, 0.0096] & 0.964 [0.950, 0.978] & $2.12\times 10^{-67}$ \\
Peroneus brevis & 0.0035 [0.0020, 0.0049] & 0.614 [0.439, 0.732] & $1.56\times 10^{-13}$ \\
Peroneus longus & 0.0105 [0.0072, 0.0144] & 0.771 [0.736, 0.852] & $2.17\times 10^{-24}$ \\
Rectus femoris & 0.0216 [0.0194, 0.0239] & 0.929 [0.899, 0.955] & $1.96\times 10^{-51}$ \\
Semitendinosus & $-0.0020$ [$-0.0027$, $-0.0013$] & $-0.692$ [$-0.786$, $-0.574$] & $5.24\times 10^{-18}$ \\
Soleus & 0.0078 [0.0067, 0.0091] & 0.929 [0.908, 0.979] & $1.95\times 10^{-51}$ \\
Tibialis anterior & 0.0014 [$-0.0059$, 0.0075] & 0.083 [$-0.298$, 0.535] & $0.371$ \\
Vastus lateralis & 0.0135 [0.0125, 0.0145] & 0.903 [0.858, 0.951] & $4.91\times 10^{-44}$ \\
Vastus medialis & 0.0231 [0.0215, 0.0245] & 0.945 [0.918, 0.969] & $2.55\times 10^{-57}$ \\
\midrule
\multicolumn{4}{@{}l}{\textbf{b. Regional scale}} \\
Biceps femoris, long head & 0.0094 [0.0083, 0.0105] & 0.916 [0.874, 0.952] & $2.22\times 10^{-47}$ \\
Gastrocnemius lateralis & 0.0021 [0.0017, 0.0025] & 0.801 [0.717, 0.867] & $1.87\times 10^{-27}$ \\
Gastrocnemius medialis & 0.0068 [0.0062, 0.0076] & 0.900 [0.859, 0.929] & $1.71\times 10^{-43}$ \\
Peroneus brevis & 0.0065 [0.0051, 0.0080] & 0.704 [0.595, 0.810] & $6.86\times 10^{-19}$ \\
Peroneus longus & 0.0103 [0.0079, 0.0125] & 0.669 [0.543, 0.763] & $1.35\times 10^{-16}$ \\
Rectus femoris & 0.0150 [0.0127, 0.0170] & 0.903 [0.853, 0.945] & $4.71\times 10^{-44}$ \\
Semitendinosus & 0.0002 [$-0.0006$, 0.0011] & 0.050 [$-0.166$, 0.299] & $0.591$ \\
Soleus & 0.0067 [0.0061, 0.0073] & 0.905 [0.865, 0.933] & $2.57\times 10^{-44}$ \\
Tibialis anterior & 0.0104 [0.0077, 0.0137] & 0.779 [0.670, 0.887] & $4.03\times 10^{-25}$ \\
Vastus lateralis & 0.0157 [0.0142, 0.0171] & 0.923 [0.887, 0.951] & $3.87\times 10^{-49}$ \\
Vastus medialis & 0.0185 [0.0167, 0.0202] & 0.933 [0.899, 0.959] & $2.26\times 10^{-52}$ \\
\bottomrule
\end{tabular*}
\par\smallskip\raggedright\footnotesize
$n=120$ per condition, adjusted for recorded height and simulated walking speed. Brackets show 95\% percentile bootstrap confidence intervals (CIs; 10,000 subject resamples, seed 42); adjusted $P$ values use Benjamini--Hochberg correction across 11 muscles. Soleus contributes once.
\end{table}

\section*{Supplementary Movies}
\begin{itemize}
\item \textbf{Supplementary Movie 1. Overview.} Demonstration of the MyoAcross framework, illustrating cross-individual muscle control and policy reuse across models constructed from different input modalities.
\item \textbf{Supplementary Movie 2. Policy performance and input sensitivity.} Comparisons of policy variants and virtual model augmentation, with demonstrations of control under unilateral lower-limb shortening and input perturbations.
\item \textbf{Supplementary Movie 3. Biomechanical validation.} Comparisons of simulated and measured vertical ground reaction forces, and simulated muscle activations with measured sEMG, across Gait120 subjects.
\end{itemize}

\bibliographystyle{IEEEtran}
\bibliography{references}

\end{document}